\documentclass[dvipsnames,format=sigconf,anonymous=false]{acmart}
\usepackage{multirow}
\usepackage{graphicx}
\usepackage{colortbl}
\usepackage{xcolor}
\definecolor{good}{HTML}{FAE8EB} 
\definecolor{bad}{HTML}{ECF4D4} 
\usepackage{bm}
\usepackage{amsmath}
\usepackage{pifont}
\usepackage{subfigure}
\usepackage{graphicx}
\usepackage{hyperref}
\usepackage{latexsym}
\usepackage{enumerate}
\usepackage{multirow}
\hypersetup{
    colorlinks=true,
    linkcolor=blue,
    filecolor=blue,      
    urlcolor=blue,
    citecolor=cyan,
}
\usepackage[linesnumbered, ruled]{algorithm2e}
\usepackage{color}
\newtheorem{myDef}{\bf Definition}
\newcommand{\post}[1]{\gdef\post{#1}}

\AtBeginDocument{%
  \providecommand\BibTeX{{%
    \normalfont B\kern-0.5em{\scshape i\kern-0.25em b}\kern-0.8em\TeX}}}

\copyrightyear{2024}
\acmYear{2024}
\setcopyright{acmlicensed}\acmConference[GECCO '24]{Genetic and Evolutionary Computation Conference}{July 14--18, 2024}{Melbourne, VIC, Australia}
\acmBooktitle{Genetic and Evolutionary Computation Conference (GECCO '24), July 14--18, 2024, Melbourne, VIC, Australia}
\acmDOI{10.1145/3638529.3654024}
\acmISBN{979-8-4007-0494-9/24/07}

\begin{document}
\author{Rongguang Ye}
\affiliation{%
  \institution{Guangdong Provincial Key Laboratory of Brain-inspired Intelligent Computation\\ Department of Computer Science and Engineering\\ Southern University of Science and Technology}
  \city{Shenzhen}
  \country{China}}
\email{yerg2023@mail.sustech.edu.cn}

\author{Longcan Chen}
\affiliation{%
  \institution{Guangdong Provincial Key Laboratory of Brain-inspired Intelligent Computation \\
  Department of Computer Science and Engineering\\
  Southern University of Science and Technology}
  \city{Shenzhen}
  \country{China}}
  
  \email{12250061@mail.sustech.edu.cn}

\author{Jinyuan Zhang\textsuperscript{*}}
\affiliation{%
 \institution{Guangdong Provincial Key Laboratory of Brain-inspired Intelligent Computation\\ Department of Computer Science and Engineering\\ Southern University of Science and Technology}
  \city{Shenzhen}
  \country{China}}
  \email{zhangjy@sustech.edu.cn}

\author{Hisao Ishibuchi\textsuperscript{*}}
\thanks{* Corresponding author.}
\affiliation{%
  \institution{Guangdong Provincial Key Laboratory of Brain-inspired Intelligent Computation\\ Department of Computer Science and Engineering\\ Southern University of Science and Technology}
  \city{Shenzhen}
  \country{China}}
  \email{hisao@sustech.edu.cn}

\title{Evolutionary Preference Sampling for Pareto Set Learning}

\renewcommand{\shortauthors}{Rongguang Ye and Longcan Chen, et al.}

\begin{abstract}
Recently, Pareto Set Learning (PSL) has been proposed for learning the entire Pareto set using a neural network. PSL employs preference vectors to scalarize multiple objectives, facilitating the learning of mappings from preference vectors to specific Pareto optimal solutions. Previous PSL methods have shown their effectiveness in solving artificial multi-objective optimization problems (MOPs) with uniform preference vector sampling. The quality of the learned Pareto set is influenced by the sampling strategy of the preference vector, and the sampling of the preference vector needs to be decided based on the Pareto front shape. However, a fixed preference sampling strategy cannot simultaneously adapt the Pareto front of multiple MOPs. To address this limitation, this paper proposes an Evolutionary Preference Sampling (EPS) strategy to efficiently sample preference vectors. Inspired by evolutionary algorithms, we consider preference sampling as an evolutionary process to generate preference vectors for neural network training. We integrate the EPS strategy into five advanced PSL methods. Extensive experiments demonstrate that our proposed method has a faster convergence speed than baseline algorithms on 7 testing problems. Our implementation is available at \href{https://github.com/rG223/EPS}{\texttt{https://github.com/rG223/EPS}}.
\end{abstract}




\maketitle
\section{Introduction}
Multi-objective optimization problems (MOPs) widely exist in real-world applications. For example, in scenarios like engineering design, it is essential to simultaneously optimize objectives such as efficient aerodynamic performance and structural robustness \cite{rai2003modular,sinha2018pareto,zhou2022multi}. However, the objectives in MOPs often conflict \cite{miettinen1999nonlinear}. Over the past decades, there has been rapid progress in developing multi-objective optimization algorithms. Among them, multi-objective evolutionary algorithms (MOEAs) have gained much attention \cite{zitzler1999evolutionary,deb2011multi,zhou2011multiobjective}. MOEAs aim to find solutions with different trade-offs that are mutually non-dominated \cite{ehrgott2005multicriteria,sener2018multi,mahapatra2020multi,ruchte2021scalable}, commonly referred to as Pareto optimal solutions. The collection of all Pareto optimal solutions is known as the Pareto set, and the corresponding objective values form the Pareto front.

\begin{figure}[t]
    \centering
    \setlength{\abovecaptionskip}{0.2cm}
    \includegraphics[width=0.99\linewidth]{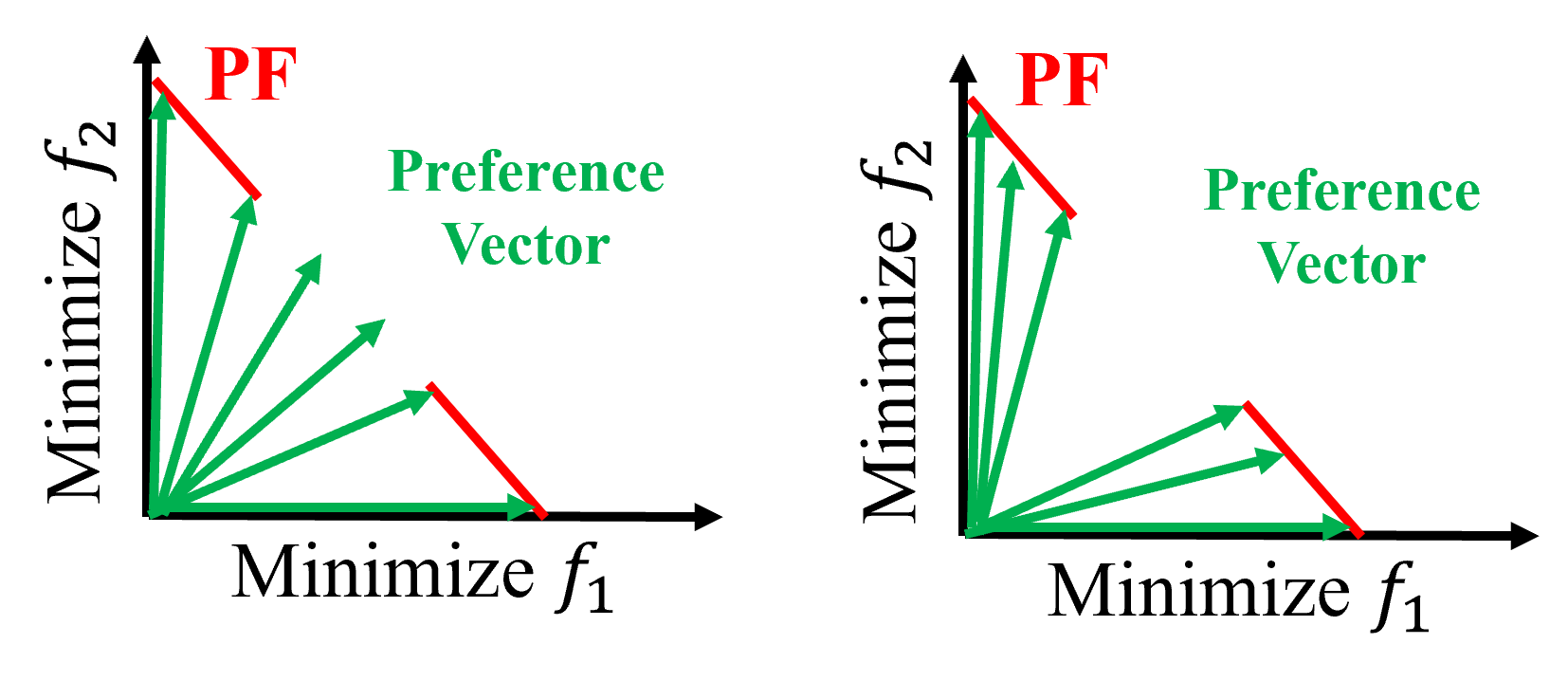}
    \caption{The sampling distribution of preference vector. \textbf{Left figure:} uniform distribution. \textbf{Right figure:} sampling preference vector focuses on the location of the Pareto front (PF).}
    \label{fig:motiv}
    \vspace{-10pt}
\end{figure}

Previously proposed MOEAs have demonstrated effectiveness in obtaining well-distributed solutions to solving various complex problems \cite{kalyanmoy2002fast,zhang2007moea,ma2021adaptive}. However, these methods still face the challenge of fully representing the entire Pareto front. This limitation arises from the finite number of solutions MOEAs can provide. One of the challenges of MOPs is to obtain an estimation of the entire Pareto front. Some researchers address this by fitting a model, such as a hyperplane or polynomial function, to approximate a continuous Pareto front using solutions in the objective space \cite{tian2018guiding,zapotecas2014using}. While these models offer insights into the trade-offs between objectives, they fall short of providing a continuous representation of the Pareto set.

To obtain a continuous Pareto set, Pareto Set Learning (PSL) has emerged as an approach to learning the complete Pareto set, concurrently obtaining the entire Pareto front \cite{lin2022pareto,lin2022pareton,navon2020learning,zhang2023hypervolume}. PSL employs a neural network to learn the mapping between preference vectors and corresponding optimal solutions. During the training phase, preference vectors are sampled from a specified distribution. The neural network output solutions according to preference vectors. The neural network then evaluates the solution and plugs evaluated solutions into the scalarization function as its loss function. Post-training, the neural network generates a corresponding solution given a preference vector. PSL allows for obtaining a complete continuous Pareto front.

Many PSL algorithms have demonstrated superior performance on benchmark problems. However, there are better preference sampling strategies than relying solely on a fixed sampling method when dealing with more complex problems. In real-world applications, the shape and location of the Pareto front are often unknown. This is because the effectiveness of preference sampling strategies depends on the Pareto front shape. Specific preference vectors are effective over others in model training, particularly when considering the location of the Pareto front. As illustrated in Figure \ref{fig:motiv}, current PSL algorithms sample preference vectors uniformly. However, some vectors within this uniform sampling may be redundant. To address this issue, as shown in the right figure of Figure \ref{fig:motiv}, we design an effective strategy that involves adaptively adjusting the preference vector sampling. This adaptive approach has the potential to enhance the neural network's performance by focusing on crucial regions.

To enhance the learning efficiency of neural networks, we propose an Evolutionary Preference Vector Sampling (EPS) strategy. Initially, uniform sampling is employed in the first period, collecting preference vectors along with their corresponding fitness values. Subsequently, we select a subset of the preference vectors as the initial population. The initial population undergoes crossover and mutation to generate preference vectors for the next period. Continuous evolution is facilitated by updating the population through subset selection in each period. This approach guarantees the provision of superior preference vectors to the neural network, thereby accelerating its convergence. Our main contributions can be summarized as follows:

\begin{itemize}

\item  We experimentally reveal the significant impact of different sampling strategies on the convergence speed of neural networks.
    
\item  We propose a novel Evolutionary Preference Vector Sampling (EPS) strategy, leveraging insights from evolutionary algorithms to sample preference vectors.

\item  We experimentally compare our proposed EPS strategy with a typical uniform sampling method on three benchmarks and four real-world problems. Five state-of-the-art algorithms are used as the baseline for comparison. Results demonstrate that EPS enhances the convergence speed of baselines on most problems.

\end{itemize}

\section{Background}

\subsection{Multi-Objective Optimization}
A multi-objective optimization problems (MOP) with $m$ objectives and $d$ decision variables can be defined as follows:

\begin{equation}
    \min_{\bm{x}\in\mathcal{X}\subset\mathbb{R}^d}\bm{f}(\bm{x})=(f_1(\bm{x}),f_2(\bm{x}),...,f_m(\bm{x})),
\end{equation}
where $\bm{x}$ is the decision variables, $\mathcal{X}$ is the decision space and $\bm{f}(\bm{x})$ is the objective functions: $\mathbb{R}^{d} \rightarrow \mathbb{R}^{m}$. No single solution can simultaneously optimize all objectives in MOPs. Usually, different solutions are compared using the Pareto dominance relationship. The definition of Pareto dominance is shown as follows: 

\begin{myDef} \textbf{(Pareto Dominance).}
  Solution $\bm{x}_{1}$ dominates solution $\bm{x}_{2}$, denoted as $\bm{x}_{1}\prec \bm{x}_{2}$, if and only if $\bm{f}_{i}(\bm{x}_{1}) \leq  \bm{f}_{i}(x_{2}), \forall i \in \{1,\ldots,m\}$ and $\exists~ j \in \{1,...,m\}$ such that $\bm{f}_{j}(\bm{x}_{1}) < \bm{f}_{j}(\bm{x}_{2})$.
\end{myDef}

Solution $\bm{x}_{1}$ is called the Pareto optimal solution if there does not exist a solution in $\mathcal{X}$ that dominates $\bm{x}_{1}$.

\begin{myDef} \textbf{(Pareto Set and Pareto Front).}
The set of all Pareto optimal solutions is called \textit{Pareto set} (denoted as $\mathcal{M}_{ps}$). The \textit{Pareto front} ($\mathcal{P}(\mathcal{M}_{ps})$) is the image of the Pareto set in the objective space.
 \end{myDef}

MOPs aim to find a set of well-distributed Pareto optimal solutions on the Pareto set.

\begin{figure}[t]
    \centering
    \includegraphics[width=0.99\linewidth]{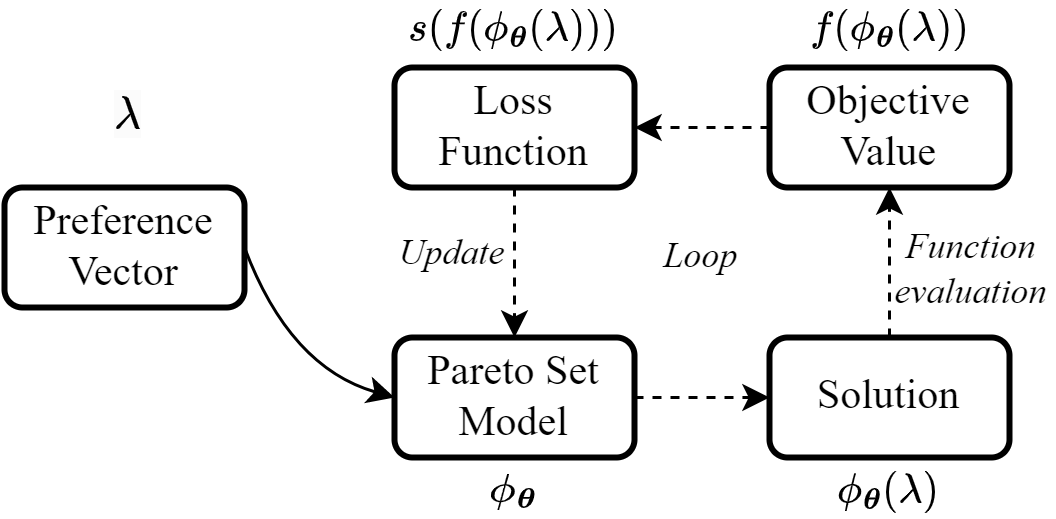}
    \caption{The process of Pareto set learning.}
    \label{fig:psl}
\end{figure}

\begin{figure*}[t]
    \centering
    \includegraphics[width=0.84\linewidth]{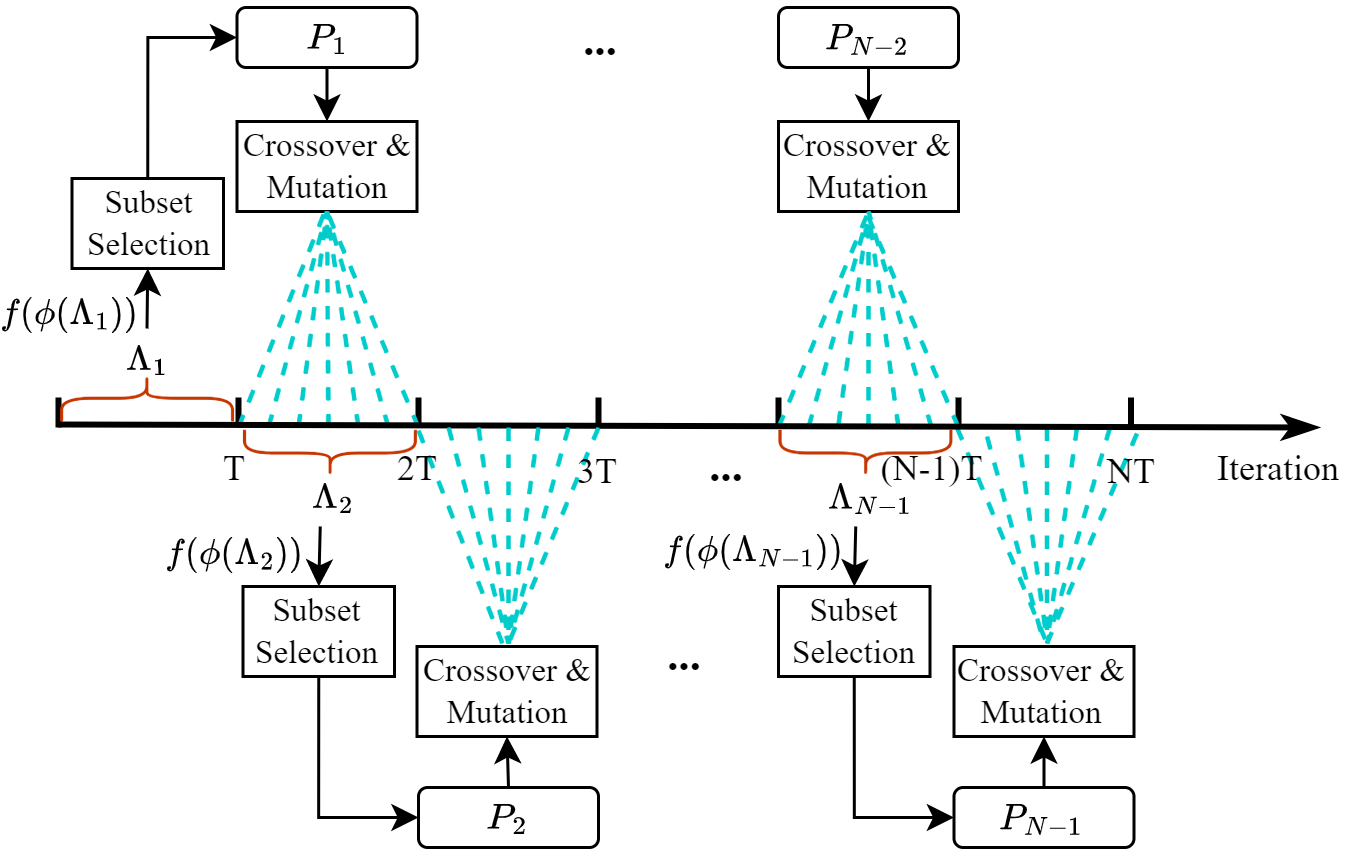}
    \caption{Evolutionary preference vector sampling strategies. For every $T$ iterations, we will collect the preference vectors $\Lambda_{i} (i=1,...N-1)$ during the period (0-th to T-th iterations collect uniformly sampled preference vectors). Afterward, subset selection is executed on these preference vectors to select outstanding individuals. Then, the population $P$, which is composed of these individuals, provides preference vectors for the next period through crossover and mutation.}
    \label{fig:epss}
\end{figure*}

\subsection{Scalarization Functions}

Given a set of preference vectors $\Lambda =\{\bm{\lambda} \in {\mathbb{R}_{+}^m} | \sum_{i=1}^m\lambda_i=1\}$, a scalarization function transforms a MOP into a single objective optimization problem. The goal of each preference vector is to obtain a Pareto optimal solution. The most commonly used scalarization function is the linear aggregation (also called the weighted sum):

\begin{equation}
    s_{ls} (\bm{x}|\bm{\lambda})=\sum_{i=1}^m\lambda_if_i(\boldsymbol{x}).
\end{equation}

The effect of the linear aggregation is limited (only works for the convex Pareto front \cite{boyd2004convex}). In contrast, the Tchebycheff aggregation function is a more comprehensive method because it can theoretically be applied to any shape of the Pareto front. Its formula is as follows:
\begin{equation}
    s_{tch}(\boldsymbol{x}|\boldsymbol{\lambda})=\max_{1\leq i\leq m}\{\lambda_i(\bm{f}_i(\boldsymbol{x})-(z_i^*-\varepsilon))\},
\end{equation}
where $z_i^*=\min_{\boldsymbol{x}\in\mathcal{X}}f_i(\boldsymbol{x})$ is an ideal value for each objective, and $\varepsilon \in \mathbb{R}_{+}$ is a small value. 

Tchebycheff aggregation function obtains the solution in the objective space that deviates from the direction of the preference vector. Therefore, the modified Tchebycheff aggregation function was proposed to obtain the solution at the intersection of the preference vector and the Pareto front \cite{wang2020survey}. Its formula is as follows:
\begin{equation}
    s_{mtch}(\boldsymbol{x}|\boldsymbol{\lambda})=\max_{1\leq i\leq m}\{{\frac{1}{\lambda_{i}}}(\bm{f}_i(\boldsymbol{x})-(z_i^*-\varepsilon))\}.
\end{equation}

\section{Evolutionary Preference Sampling}
Preference vectors are often sampled from a uniform distribution in previous works on Pareto set learning. Improving the efficiency of sampling can accelerate the convergence speed of the Pareto set model, so we propose a novel sampling strategy to achieve it.

\subsection{Pareto Set Model}

A Pareto set model can be defined as follows:

\begin{equation}
    \bm{x} = \phi_{\bm{\theta}}(\bm{\lambda}),
\end{equation}
where $\bm{\lambda} \in \mathbb{R}^{m}$ is the preference vector, $\bm{\theta}$ is the parameter of the Pareto set model, and $\phi_{\bm{\theta}}: \mathbb{R}^{m} \rightarrow \mathbb{R}^{d}$. As shown in Figure \ref{fig:psl}, the Pareto set model ($\phi_{\bm{\theta}}$) fits the mapping relationship from the preference vector $\bm{\lambda}$ to the Pareto optimal solution during training. First, the solution generated by the Pareto set model is evaluated through the objective functions. Then, using the scalarization function optimizes the Pareto set model. The task of optimizing the Pareto set model is formulated as follows:

\begin{equation} \bm{\theta}^{*} = \arg\min_{\bm{\theta}}\mathbb{E}_{{\bm{\lambda}}\in \mathcal{D}} 
 [s(f(\phi_{\bm{\theta}}(\bm{\lambda}))],
\end{equation}
where $\mathcal{D}$ indicates a predefined distribution. After training, the Pareto set model can provide a solution for any preference of the decision-maker.

\subsection{Evaluation of Preference Vectors}
In this subsection, we will consider how to evaluate the preference vectors because the preference vectors are regarded as optimization variables. For a given preference vector set $\Lambda$ and a Pareto set model $\phi_{\bm{\theta}}$, we can first use $\phi_{\bm{\theta}}$ to calculate the solution set $X=\phi_{\bm{\theta}}(\Lambda)$. Afterward, the objective functions evaluate the solution set to obtain the objective values $\bm{f}(X)$. At this time, we define the evaluated function $g(\bm{\lambda})$ of the preference vector $\bm{\lambda}$ as follows:

\begin{equation} \label{eva}
    g(\bm{\lambda}) = \bm{f}(\phi_{\bm{\theta}}(\bm{\lambda})).
\end{equation}

In other words, the evaluated function $g(\cdot)$ depends on the performance of the Pareto set model. The accuracy of the Pareto set model is not ideal in the early stages of training. However, its evaluation of preference vectors will become increasingly accurate as the model is continuously optimized. In addition, the evaluation method we use does not add any computational burden because all computation has already been done in the Pareto set learning framework (Figure \ref{fig:epss}).

\subsection{Preference Subset Selection and Generation}
In this subsection, we will discuss how to select individuals forming the population, then they can generate preference vectors to the Pareto set model. The preference vectors are the input of the Pareto set model, and it is crucial for the Pareto set learning to receive high-quality preference vectors in the training process. As shown in Figure \ref{fig:epss}, the preference set $\Lambda_{t}$ is collected in $t$-th period, and can be evaluated by Eq. (\ref{eva}). After evaluating the preference vector set $\Lambda_{t}$, the preference vectors generating solutions with strong convergence are considered. The goal is to select the proportion of $sp$ in $\Lambda_{t}$ as the next generation population $P_{t}$. In our work, we use \textit{Non-Dominated Sorting} \cite{kalyanmoy2002fast} and \textit{Crowding Distance} \cite{kalyanmoy2002fast} to select $sp \cdot |\Lambda_{t}|$ preference vectors in $\Lambda_{t}$.

\begin{figure}[t]
    \centering
    \includegraphics[width=0.99\linewidth]{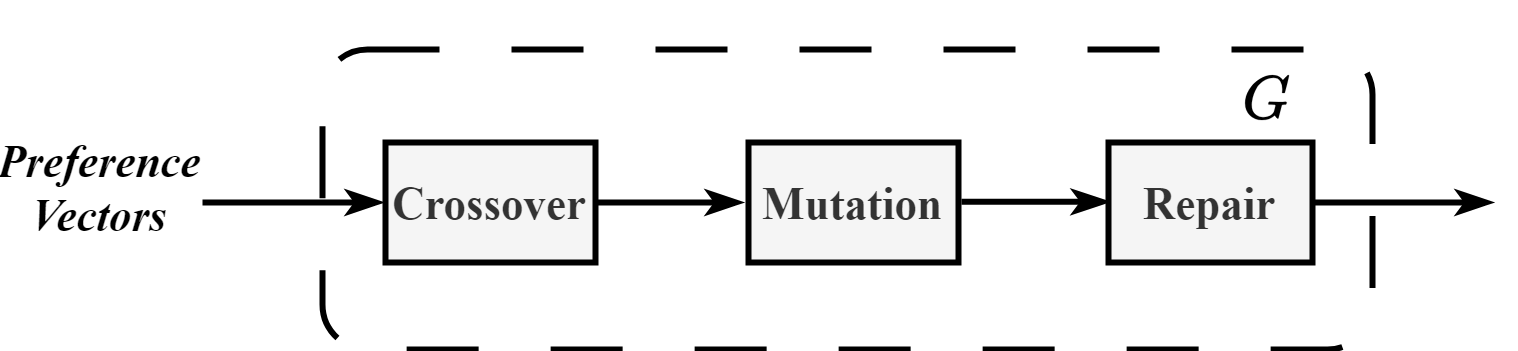}
    \caption{The process of preference vector sampling by population. This process includes three parts: crossover, mutation, and repair mechanism.}
    \label{fig:g}
\end{figure}



Subsequently, we will generate preference vectors using the collected preference vectors to train the Pareto set model in the next period. As shown in Figure \ref{fig:epss}, when $P_{t}$ is selected through subset selection, $P_{t}$ can generate preference vectors for the next period (from $t+1$ to $t+T$ iteration) through simulated binary crossover (SBX) \cite{deb1995simulated} and polynomial mutation (PM) \cite{deb2007self}. Specifically, we randomly select the parents $\bm{\lambda}^{(t,1)}$ and $\bm{\lambda}^{(t,2)}$ from $P_{t}$, then the children $\bm{\lambda}^{'(t, 1)}$ and $\bm{\lambda}^{'(t, 2)}$ will be generated.



 
Besides, it's possible for the sum of preference vectors generated by crossover and mutation to exceed one. Therefore, the sum of $\bm{\lambda}^{''(t, i)}$ should be repaired to one:

\begin{equation}
\bm{\lambda}^{''(t,i)}= 
    \frac{\bm{\lambda}^{'(t,i)}}{\sum_{j=1}^{m} \bm{\lambda}_{j}^{'(t,i)}}.
\end{equation}

Figure \ref{fig:g} shows the flowchart of generating preference vectors based on population. The preference vector generator ($G$) includes the three parts. We use $G$ to generate preference vectors for the next period. Our optimization problem for the next period can be written as the following formula:

\begin{equation}
    \min_{\bm{\theta}}\mathbb{E}_{\bm{\lambda}\sim G(P_t)}s(f(\phi(\bm{\lambda}))),
\end{equation}
where $CM(P_{t})$ represents the crossover and mutation operator in population $P_{t}$ by random sample parents. In addition, the population is continually evolving every period in the training process. It's beneficial for the Pareto set model to accelerate the convergence speed. The process of the evolutionary preference sampling is also summarized in Algorithm \ref{alg:eps}.

\begin{algorithm}[t]
  \KwIn{maximum iterations $T_{\text{max}} = NT$, Pareto set model $\phi_{\bm{\theta}_{0}}$, learning rate $\alpha$, objective functions $\bm{f}$, and scalarization function $s$.}
  Initialize preference set $P=\emptyset$\;
  \For{$t = 1$ \KwTo $T$}{
    $\Lambda_{t} \sim \mathcal{U}(0, 1)$ $\rhd$ sample $\Lambda_{t}$ from uniform distribution\;
    $X = \phi_{\bm{\theta}_{t-1}}(\Lambda_{t})$ $\rhd$ Pareto set model output solution\;
    $Y = \bm{f}(X)$ $\rhd$ Function evaluation\;
    $P_{0} = P_{0} \cup \{\Lambda_{t}: Y \}$ $\rhd$ Data collection \;
    $\bm{\theta}_{t} = \bm{\theta}_{t-1} - \alpha \frac{\partial s(Y)}{\partial \bm{\theta}}$ $\rhd$ Update the Pareto set model\;
  }
  Choose a subset $P$ from $P_{0}$ by \textit{Non-Dominated Sorting} and \textit{Crowding Distance}\;
  $P_{0} = \emptyset$\;
  \For{$t = T+1$ \KwTo $T_{max}$}{
    $\Lambda_{t} \sim CM_{P}$ $\rhd$ $P$ generate $\Lambda_{t}$ by SBX and PM\;
    $X = \phi_{\bm{\theta}_{T+t-1}}(\Lambda_{t})$ $\rhd$ Pareto set model output solution \;
    $Y = \bm{f}(X)$ $\rhd$ Function evaluation \;
    $P_{0} = P_{0} \cup \{\Lambda_{t}: Y \}$ $\rhd$ Data collection \;
    $\bm{\theta}_{T+t} = \bm{\theta}_{T+t-1} - \alpha \frac{\partial s(Y)}{\partial \bm{\theta}}$ $\rhd$ Update the Pareto set model\;
    
    \If{$t \% T = 0$}{
      Choose a subset $P$ from $P_{0}$.\;
      $P_{0} = \emptyset$\;
    }
  }
  \KwOut{$\phi$}
  \caption{Evolutionary Preference Sampling \label{alg:eps}}
\end{algorithm}

\begin{table*}[t]
\setlength{\tabcolsep}{5mm}{}
\setlength{\abovecaptionskip}{0cm}
  \caption{Mean and standard deviation of the convergence results (log HV difference) of the five baseline algorithms on seven problems with or without using the EPS strategy over 11 runs. Red and brown backgrounds indicate improvements and no improvements compared to the original algorithm respectively.}
  \label{tab}

\begin{tabular}{cclllll}
\hline
\multicolumn{1}{c}{\multirow{2}{*}{Problem}} & \multicolumn{1}{c}{\multirow{2}{*}{EPS}} & \multicolumn{5}{c}{Method}                                         \\ \cline{3-7} 
\multicolumn{1}{c}{}                         & \multicolumn{1}{c}{}                     & PSL-MTCH    & PSL-TCH     & COSMOS      & PSL-LS      & PSL-HV      \\ \hline
\multirow{2}{*}{ZDT3}                         & \ding{55}                                         & -1.58 (1.21) &  -1.04 (0.49) & -1.12 (0.18) & -0.90 (1.32) & -0.80 (0.19) \\
                                              & \ding{51}                                         & \cellcolor{good}-2.23 (1.35) & \cellcolor{good} -1.47 (1.35) & \cellcolor{good} -1.81 (0.73) & \cellcolor{good} -1.36 (1.64) & \cellcolor{good} -1.17 (0.59) \\
\multirow{2}{*}{DTLZ5}                        & \ding{55}                                         & -6.15 (0.49) & -4.19 (0.35) & -2.74 (0.21) & -4.47 (1.98) & -3.86 (0.37) \\
                                              & \ding{51}                                         & \cellcolor{good} -6.92 (0.40) & \cellcolor{good} -6.07 (0.59) & \cellcolor{good} -3.01 (0.27) & \cellcolor{good} -6.52 (0.57) & \cellcolor{good} -4.88 (1.41) \\
\multirow{2}{*}{DTLZ7}                        & \ding{55}                                         & -0.36 (0.44) & -0.20 (0.24) & -1.11 (0.11) & -0.22 (0.67) & -0.40 (0.29) \\
                                              & \ding{51}                                         & \cellcolor{good} -0.60 (0.72) & \cellcolor{good} -0.61 (0.68) & \cellcolor{good} -1.35 (0.11) & \cellcolor{good} -0.30 (0.74) & \cellcolor{good} -0.46 (0.55) \\
\multirow{2}{*}{RE21}                         & \ding{55}                                         & -0.21 (0.16) & -0.13 (0.36) & 1.26 (0.23)  & -1.12 (0.08) & -0.06 (0.33) \\
                                              & \ding{51}                                         & \cellcolor{good} -0.70 (0.08) & \cellcolor{good} -0.62 (0.19) & \cellcolor{good} 0.75 (0.24)  & \cellcolor{bad} -0.75 (0.07) & \cellcolor{bad} 0.35 (0.41)  \\
\multirow{2}{*}{RE33}                         & \ding{55}                                         & 3.44 (0.09)  & 3.43 (0.06)  & 2.91 (0.61)  & 3.69 (0.43)  & 3.53 (0.17)  \\
                                              & \ding{51}                                         & \cellcolor{good} 3.33 (0.01)  & \cellcolor{good} 3.32 (0.01)  & \cellcolor{good} 2.51 (0.39)  & \cellcolor{good} 3.32 (0.00)  & \cellcolor{good} 3.46 (0.13)  \\
\multirow{2}{*}{RE36}                         & \ding{55}                                         & -0.47 (0.51) & -1.79 (0.56) & 3.45 (0.12)  & 0.50 (0.28)  & 2.12 (0.28)  \\
                                              & \ding{51}                                         & \cellcolor{good} -0.99 (0.26) & \cellcolor{good} -2.48 (0.29) & \cellcolor{good} 2.47 (0.56)  & \cellcolor{good} -1.59 (0.31) & \cellcolor{good} 1.92 (0.43)  \\
\multirow{2}{*}{RE37}                         & \ding{55}                                         & -2.12 (0.76) & -2.84 (0.07) & -1.82 (0.07) & -1.71 (0.01) & -3.35 (0.04) \\
                                              & \ding{51}                                         & \cellcolor{good} -3.19 (0.69) & \cellcolor{good} -3.04 (0.41) & \cellcolor{good} -2.14 (0.06) & \cellcolor{good} -1.79 (0.24) & \cellcolor{good} -3.36 (0.05) \\ \hline
\end{tabular}
\end{table*}

\begin{figure*}[t]
    \centering
    \includegraphics[width=0.9\textwidth]{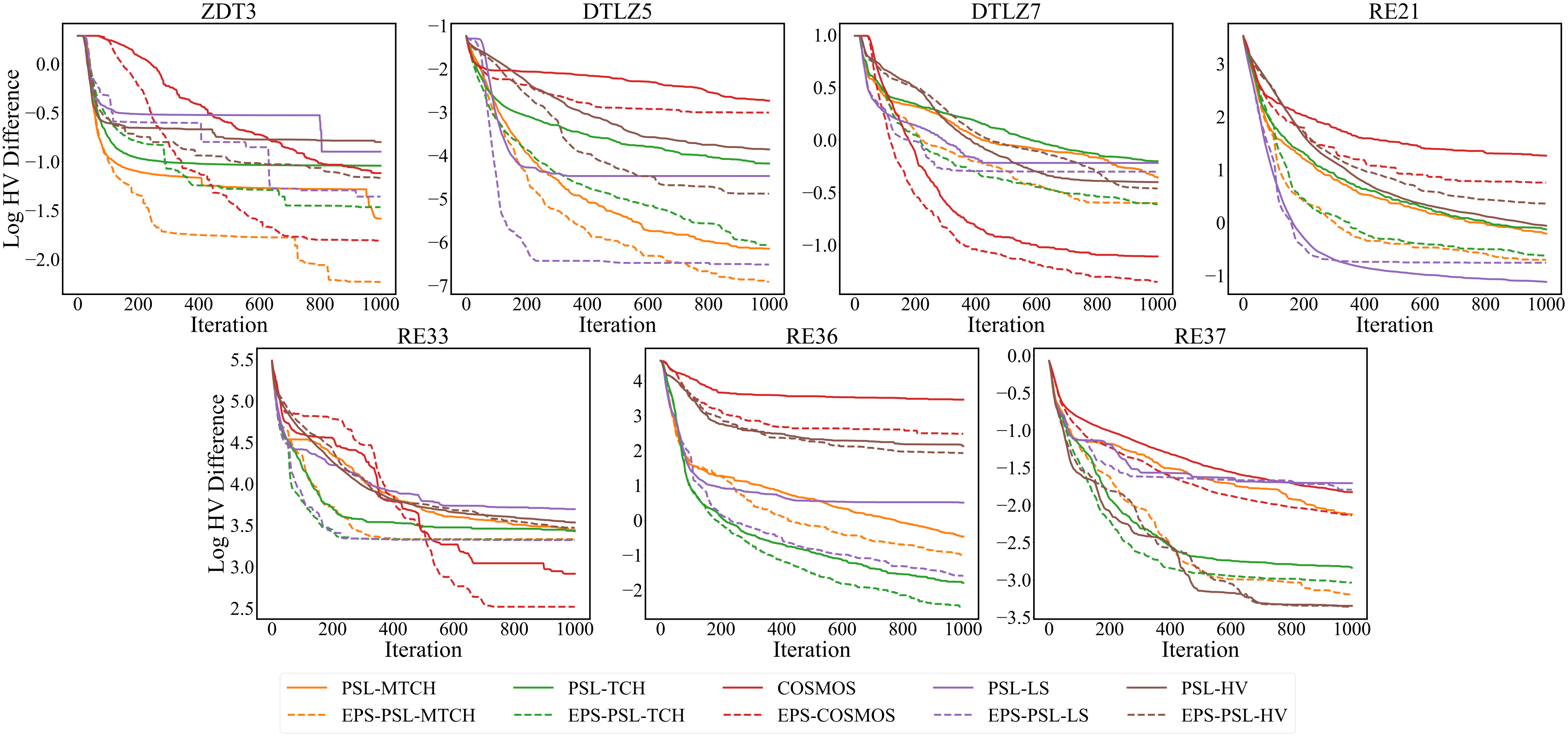}
    \caption{Convergence comparison of log HV differences on different baselines and test problems over 11 runs. Different colors indicate different algorithms, the solid line indicates that the original algorithm uses uniform sampling in preference vector sampling, and the dotted line indicates that the original algorithm uses the EPS strategy.}
    \label{fig:mainresult}
\end{figure*}

\begin{figure*}[t]
    \centering
    \includegraphics[width=0.9\linewidth]{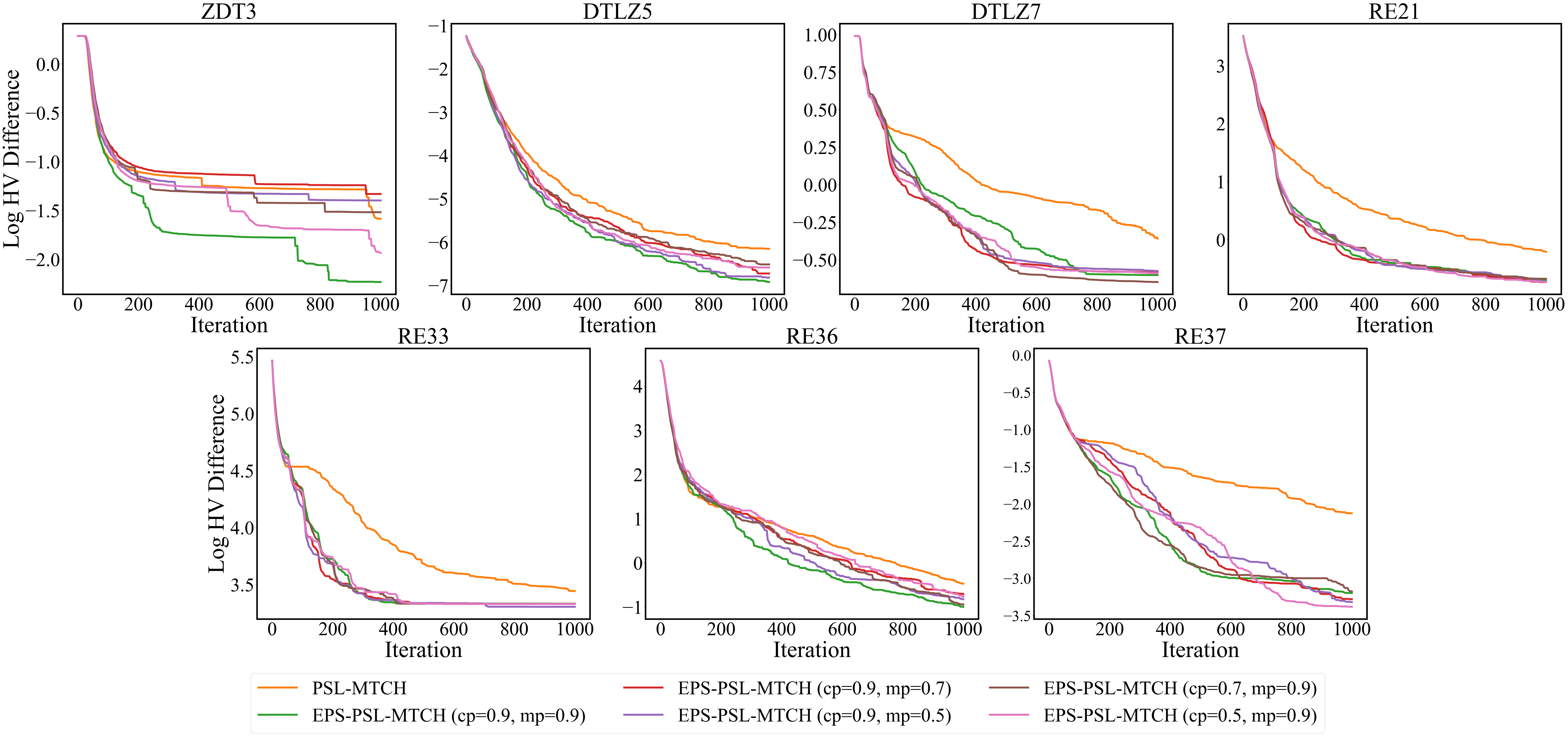}
    \caption{The sensitivity of crossover and mutation probability to EPS.}
    \label{fig:b}
\end{figure*}
\begin{figure*}[t]
    \centering
    \includegraphics[width=0.9\linewidth]{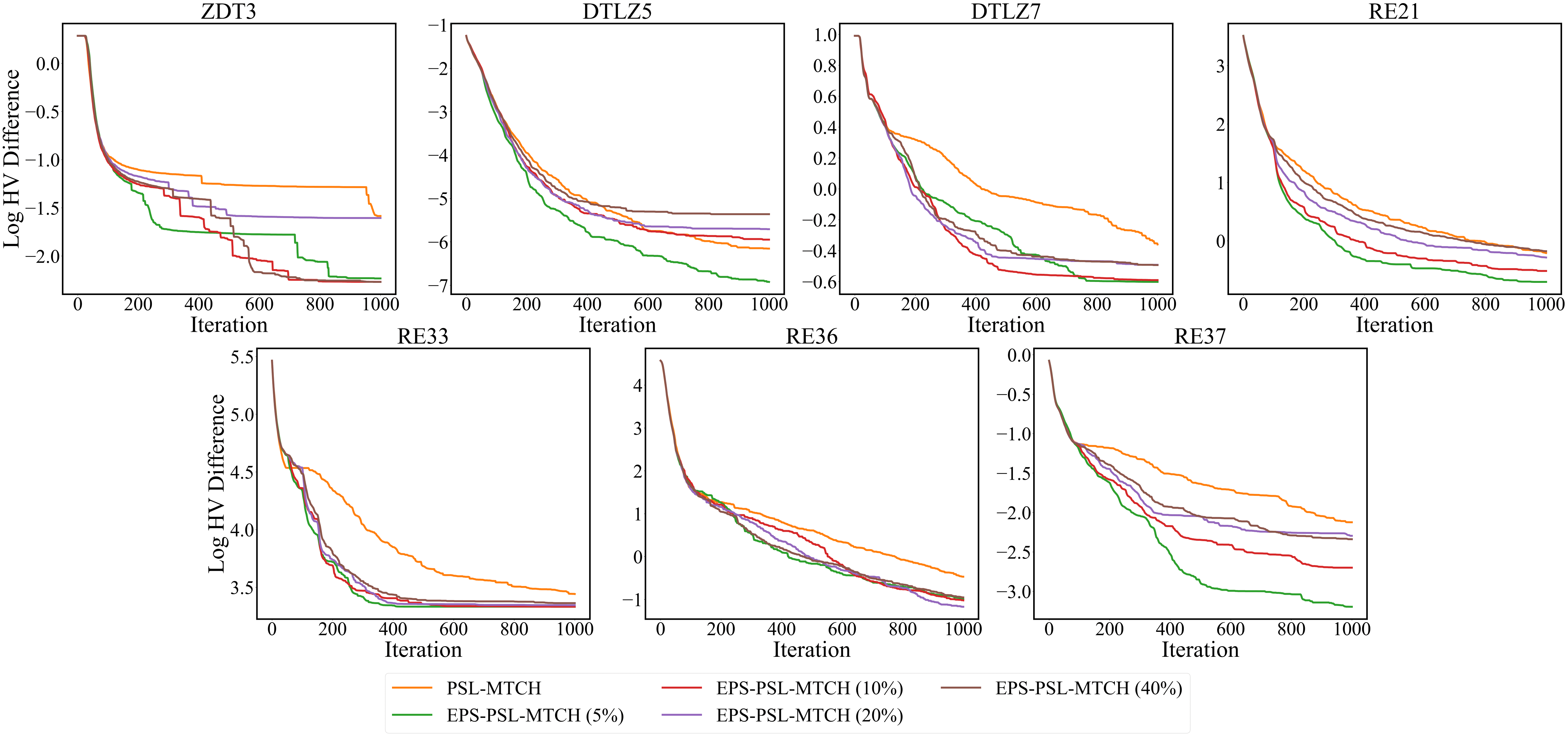}
    \caption{The sensitivity of the subset selection percentage to EPS.}
    \label{fig:p}
\end{figure*}
\section{Experiments}

\subsection{Experimental Settings}
We use the five state-of-the-art algorithms (PSL-LS \cite{navon2020learning}, PSL-TCH \cite{lin2022pareto}, PSL-MTCH \cite{lin2022pareton}, COSMOS \cite{ruchte2021scalable} and PSL-HV \cite{zhang2023hypervolume}) as baselines, which differ in their scalarization functions. To evaluate the effect of the evolutionary preference sampling (EPS) strategy, we embed EPS into these five algorithms. Then, several widely used benchmark problems and real-world (RE) problems with different shapes of the Pareto fronts and the number of objectives ($m$) are used for testing. The Pareto fronts of tested problems can be divided into four types:

\begin{itemize}
\item \textbf{Convex} \cite{tanabe2020easy}: RE21 ($m$=2, $d$=4).
\item \textbf{Disconnected} \cite{zitzler2000comparison,deb2005scalable}: DTLZ7 ($m$=3, $d$=10), ZDT3 ($m$=2, $d$=10).
\item \textbf{Degenerated} \cite{deb2005scalable,tanabe2020easy}: DTLZ5 ($m$=3, $d$=10), RE36 ($m$=3, $d$=4).
\item \textbf{Irregular} \cite{tanabe2020easy}: RE33 ($m$=3, $d$=4) and RE37 ($m$=3, $d$=4).
\end{itemize}
Hypervolume (HV) \cite{fonseca2006improved} is a commonly used indicator in multi-objective optimization problems. It can be used when the true Pareto front of the multi-objective optimization problem is unknown. The evaluation metric we use takes into account the difference between the approximated HV (calculated by testing algorithms) and the HV 
 (calculated by true Pareto front):

\begin{equation}
    \text{log HV difference} = log (HV + \epsilon - \Hat{HV}),
\end{equation}
where $\epsilon \in \mathbb{R}_{+}$ is a small value and $\Hat{HV}$ is approximated by algorithms. The smaller the difference of log HV difference, the better the performance. We test all algorithms on the 7 test problems over 11 runs, and use the same hyperparameters for all algorithms (learning rate is 0.001, the batch size is 8 and the maximum iteration is 1000.). The same number of uniform preference vectors are used to evaluate all algorithms. All experiments are performed with eight NVIDIA GeForce RTX 2080 Ti GPUs (11GB RAM).

\begin{figure*}[t]
\centering
\includegraphics[width=0.75\linewidth]{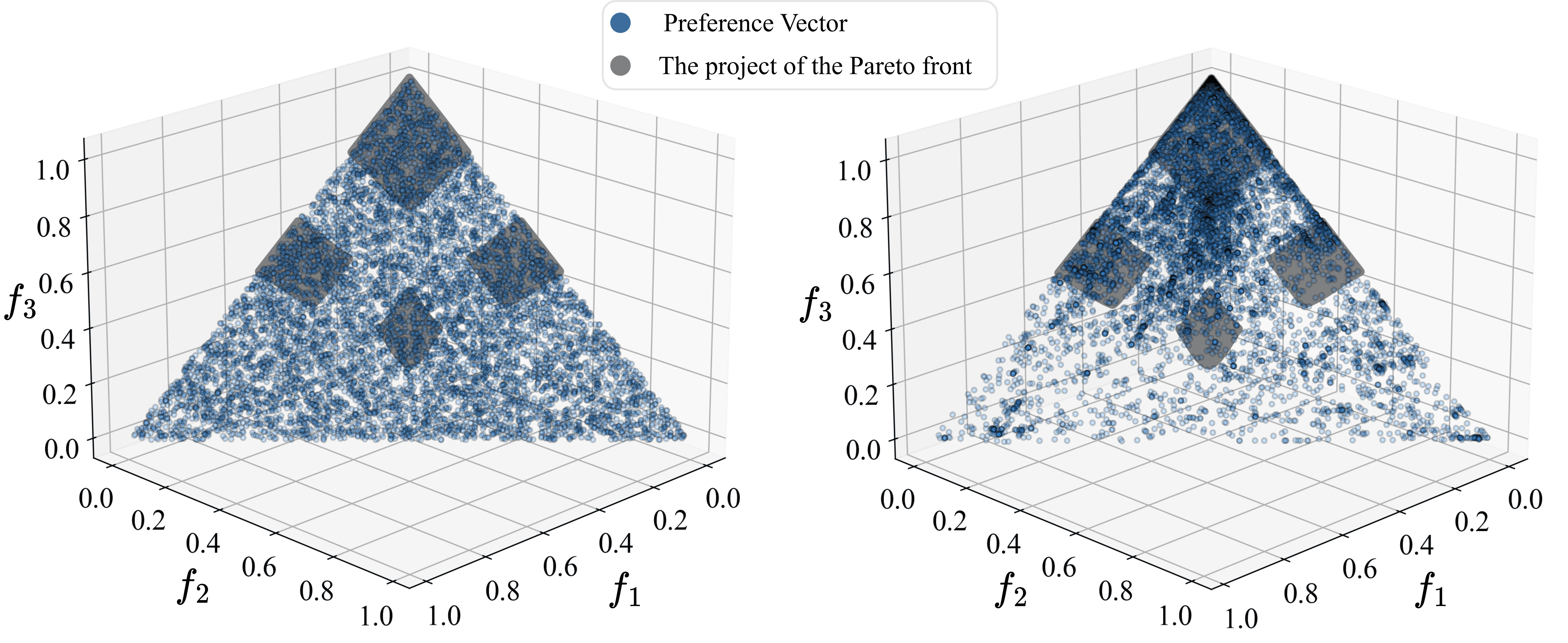} 

\caption{The preference vector distributions obtained by using uniform sampling (left) and the EPS strategy (right) for PSL-MTCH on DTLZ7. The gray shading represents the projection of the true Pareto front onto the preference vector hyperplane.} \label{ps}
\end{figure*}

\begin{figure*}[t]
\centering
\includegraphics[width=0.75\linewidth]{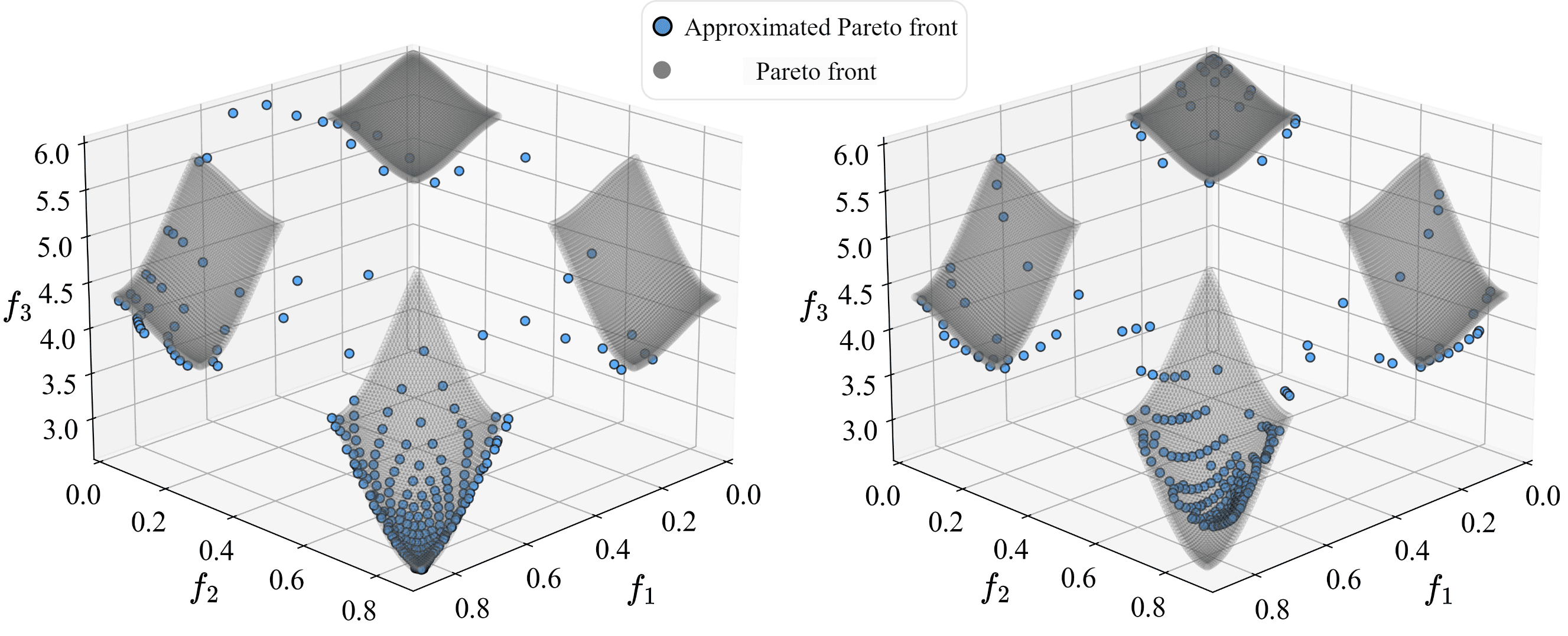} 
\caption{Pareto front obtained by PSL-MTCH using uniform sampling (left) and the EPS strategy (right) on the DTLZ7. The gray shading represents the true Pareto front.} \label{dtlz7pf}
\end{figure*}


\subsection{Main Results}
As shown in Table \ref{tab}, we show the impact of using EPS on the convergence results of various algorithms. We can find that EPS can improve the convergence result of the original algorithm in most cases. In addition, we investigate the effect of EPS on the convergence speed in each algorithm. The results in Figure \ref{fig:mainresult} show that the EPS strategy converges faster on four algorithms (PSL-LS, PSL-TCH, PSL-MTCH, and COSMOS) in most cases, but not significantly on PSL-HV. This is because PSL-HV updates the Pareto set model based on the HV maximization method, the sampling of preference vectors has limited influence on PSL-HV. The other four baselines use the scalarization function based on preference vectors (e.g., weight-sum, Tchebycheff) to update the Pareto set model. For these algorithms, the sampling of the preference vector directly impacts the optimization of the Pareto set model, so the proposed sampling method can accelerate the convergence. We can see that the effect of the proposed sampling method is significant in the ZDT3, DTLZ5, DTLZ7, and RE36 problems (comparison of dotted and solid lines of the same color). The main reason is that the shape of the Pareto front for these problems is disconnected or degenerated. The sampling efficiency of the preference vector is more important. For RE21, its Pareto front is continuous and convex in two-dimensional space. The uniform sampling preference vector in this Pareto front is also sufficient, so the use of EPS does not necessarily improve the performance. In summary, EPS can not only help the baseline algorithms improve the convergence speed in most cases, but also achieve better convergence results.

\subsection{Sensitivity Analysis}
The subset selection process that forms the next-generation population directly determines the quality of the subsequent preference vectors. To investigate the impact of the percentage ($sp$) of subset selection on the EPS strategy, we test 5\%, 10\%, 20\%, and 40\% selection percentages on PSL-MTCH and compare them with the original algorithm. The experimental results are shown in Figure \ref{fig:p}. We can find that $sp$=5\% and $sp$=10\% are more beneficial to the convergence of the Pareto set model in most problems, which means that we need high-quality individuals (solutions with strong convergence) to enter the next generation to improve the learning efficiency of the Pareto set model.

Moreover, the population relies on crossover and mutation to generate preference vectors to the Pareto set model, so we also investigate the impact of different crossover probability ($cp$) and mutation probability ($mp$) on the convergence of the Pareto set model. The results are shown in Figure \ref{fig:b}. In most problems (except ZDT3), $cp$=0.9 and $cp$=0.9 have relatively better performance, and fluctuations in crossover and mutation probability do not significantly affect the convergence speed of the algorithm. Parameter settings ($cp=0.9, mp=0.7$, and $cp=0.9, mp=0.5$) are slightly worse than the original algorithm on ZDT3. This is because the Pareto front of ZDT3 is divided into five discrete parts. Therefore, our sampling strategy needs a large enough mutation probability to jump out of the local optimum and find other parts of the Pareto front. EPS has a stable convergence speed compared with the original algorithm on the other six test problems.

\subsection{Case Study}
To further explore the advantages of the EPS strategy, we conduct a case study on the disconnected problem DTLZ7. Figure \ref{dtlz7pf} shows the Pareto front obtained by the uniform sampling and the EPS strategy respectively. It can be found that the proposed sampling strategy has a better estimation of the Pareto front than that of the uniform sampling. In particular, the distribution of solution sets obtained by uniform sampling is imbalanced. The solution sets are almost concentrated in the two sub-regions of the Pareto front, while the solution sets of the other two sub-regions are very sparse. On the contrary, the distribution of solution sets obtained by EPS is relatively balanced in the four sub-regions of the Pareto front. The left figure of Figure \ref{ps} is the distribution of preference vectors using the EPS strategy. We can find many preference vectors focused on sampling in and around the Pareto front area. However, some preference vectors are sampled in the middle of the Pareto front. It indicates the population may converge to the Pareto front, then crossover and mutate generating middle preference vectors. The right figure of Figure \ref{ps} is the distribution of preference vectors using uniform sampling. However, it's inefficient because preference vectors are globally explored. The EPS strategy is more efficient than uniform sampling in complex Pareto fronts.

\section{Related Work}
Evolutionary multi-objective algorithms (MOEAs) have rapidly developed in the past two decades to address multi-objective optimization problems. MOEAs are classified into three primary categories: decomposition-based EMOAs (e.g., MOEA/D \cite{zhang2007moea}), Pareto dominance-based MOEAs (e.g., NSGA-II \cite{kalyanmoy2002fast}), and indicator-based MOEAs (e.g., SMS-MOEA \cite{emmerich2005emo}). These approaches iteratively optimize populations to generate solution set ultimately. However, their limitation lies in obtaining only a finite set of solutions that approximate the Pareto front. 
Recent research has explored using solution set values in the objective space to construct models based on hyperplanes or polynomial functions \cite{zapotecas2014using}. While this technique can generate a continuous Pareto front, its effectiveness is constrained as it learns only the fitting equation in the objective space, without providing the corresponding Pareto set \cite{zhou2009approximating}. MORM \cite{asilian2022machine} employs an inverse model to map solution set values in the objective space to decision space values. The corresponding solutions can be derived by inputting desired objective values to the inverse model. However, determining the appropriate objective values for input poses challenges, such as unknown ranges of objective functions. A novel approach \cite{suresh2023machine} proposed by Anirudh et al. diverges from using actual solution values in the objective space. Instead, it converts these values to pseudo-weights, which are input into the model. The model subsequently suggests solutions based on the provided pseudo-weights. This method offers an alternative perspective in handling the challenge of Pareto front approximation.

Pareto set learning was recently proposed \cite{lin2022pareto} based on the neural network, which learns the mapping from preference vectors to Pareto optimal solutions by training a neural network. The advantage of this learning method is that the trained model can accept continuous preference vectors and subsequently estimate the entire Pareto set. This capability is particularly crucial in various domains, such as drug design \cite{jain2023multi} and multi-objective combinatorial optimization problems \cite{lin2022pareton}.

\section{Conclusion}
In this paper, we have shown the convergence of the Pareto set model is affected by the sampling of the preference vector. Using the uniform sampling method can not handle complicated Pareto fronts. To improve the efficiency of preference sampling, we propose an evolutionary preference sampling (EPS) strategy. At the beginning, EPS collects the preference vector of uniform sampling in the first period and the corresponding objective values. Later, we use the subset selection in the collected data to choose the first population. We use the current population through crossover and mutation to generate preference vectors for the Pareto set model in the next period. Then, the population is updated each period to evolve the preference vectors continuously. In this way, the Pareto set model can be learned effectively. We embed EPS into five algorithms in three benchmarks and four real-world problems. The experimental results show EPS can accelerate the convergence of the model in most cases.

The proposed EPS strategy has shown its effectiveness in improving the convergence of the Pareto set model in some problems. However, EPS still cannot learn the Pareto set effectively in many aspects. For example, Pareto set learning still needs satisfactory performance in multi-objective integer optimization and multi-objective problems with constraints. Further investigation on expanding EPS to more complex multi-objective optimization problems is an interesting future research topic.

\begin{acks}
This work was supported by National Natural Science Foundation of China (Grant No. 62106099, 62250710163, 62376115), Guangdong Basic and Applied Basic Research Foundation (Grant No. 2023A1515011380), Guangdong Provincial Key Laboratory (Grant No. 2020B121201001).
\end{acks}

\bibliographystyle{ACM-Reference-Format}
\bibliography{acmart}










\end{document}